\documentclass{article}

\usepackage{arxiv}
\usepackage[utf8]{inputenc} 
\usepackage[T1]{fontenc}    
\usepackage{hyperref}       
\usepackage{url}            
\usepackage{booktabs}       
\usepackage{amsfonts}       
\usepackage{nicefrac}       
\usepackage{microtype}      
\usepackage{graphicx}
\graphicspath{{./figures/}}
\usepackage{amsmath}
\usepackage{tabularx}
\usepackage{subcaption}
\usepackage{float}
\usepackage{comment}
\usepackage[numbers]{natbib}

\title{Navigation in a simplified Urban Flow through Deep Reinforcement Learning}

\author{
  Federica Tonti \\
  FLOW, Engineering Mechanics, \\
  KTH Royal Institute of Technology, \\
  Osquars backe 18, 11428, Stockholm, Sweden \\
  \texttt{ftonti@kth.se} \\
  \And
  Jean Rabault \\
  Independent Researcher, \\
  0854 Oslo, Norway \\
  \texttt{jean.rblt@gmail.com} \\
  \And
  Ricardo Vinuesa \\
  FLOW, Engineering Mechanics,\\
  KTH Royal Institute of Technology, \\
  Osquars backe 18, 11428, Stockholm, Sweden \\
  \texttt{rvinuesa@mech.kth.se}
}

\begin{document}
\maketitle

\begin{abstract}
The increasing number of unmanned aerial vehicles (UAVs) in urban environments requires a strategy to minimize their environmental impact, both in terms of energy efficiency and noise reduction. In order to reduce these concerns, novel strategies for developing prediction models and optimization of flight planning, for instance through deep reinforcement learning (DRL), are needed. 
Our goal is to develop DRL algorithms capable of enabling the autonomous navigation of UAVs in urban environments, taking into account the presence of buildings and other UAVs, optimizing the trajectories in order to reduce both energetic consumption and noise. This is achieved using fluid-flow simulations which represent the environment in which UAVs navigate and training the UAV as an agent interacting with an urban environment.
In this work, we consider a domain domain represented by a two-dimensional flow field with obstacles, ideally representing buildings, extracted from a three-dimensional high-fidelity numerical simulation. The presented methodology, using PPO+LSTM cells, was validated by reproducing a simple but fundamental problem in navigation, namely the Zermelo's problem, which deals with a vessel navigating in a turbulent flow, travelling from a starting point to a target location, optimizing the trajectory. The current method shows a significant improvement with respect to both a simple PPO and a TD3 algorithm, with a  success rate (SR) of the PPO+LSTM trained policy of 98.7\%, and a crash rate (CR) of 0.1\%, outperforming both PPO (SR = 75.6\%, CR=18.6\%) and TD3 (SR=77.4\% and CR=14.5\%).
This is the first step towards DRL strategies which will guide UAVs in a three-dimensional flow field using real-time signals, making the navigation efficient in terms of flight time and avoiding damages to the vehicle.   
\end{abstract}

\section{Introduction}
\label{introduction}

The presence of unmanned aerial vehicles (UAVs) is constantly increasing in urban environments due to the variety of tasks they can accomplish, from package delivery to surveillance and traffic monitoring, with the advantage of being able to access areas which would be difficult to reach by ground transportation or using bigger aerial vehicles, such as helicopters (\cite{Hwang2020, website1, website2}). On the other hand, the increasing number of UAVs brings new challenges which have not been faced until now because of their relatively recent use in cities, such as acoustic pollution or increasing risk of accidents (\cite{Qian2021, Torija2020, patents, Sun2024, Svaty2022}). Due to these challenges, developing an efficient strategy to allow UAVs to navigate autonomously in complex environments is becoming crucial, not only to accomplish the aforementioned tasks but also to satisfy security constraints and ideally reduce their environmental impact, in particular acoustic pollution. Under this requirements, path planning , i.e. being able to find an optimal path between a starting point and a target point, avoiding obstacles if present, becomes essential. \\
For UAV navigation problems, path planning, obstacle detection and avoidance methods can typically be divided into non-learning-based and learning-based methods. Non-learning-based methods need a good knowledge and understanding of the problem domain, a fact that leads to the difficulty to generalize to unseen environments, but they provide a good interpretability of the process since the decision-making task is based on well-defined algorithms (\cite{REDA2024}). Dijkstra's Algorithm (\cite{Dijkstra1959}, \cite{Jabbar2023}), A\textsuperscript{*}, which can be seen as an extension of the Dijkstra's Algorithm (\cite{You2023, FOEAD2021507}), or rapidly exploring random tree (RRT) (\cite{LaValle1998RapidlyexploringRT, Zhang9849614, Noreen2016}) are popular non-learning-based path planning algorithms which have demonstrated to be successful in environments which do not exhibit uncertainties, but give poor performance when the environment is dynamic. When dealing with obstacles, sensing and avoidance methods steer the vehicle in the opposite direction with respect to the obstacles and navigate through the environment by path-planning algorithms (\cite{Chandran2023}).  Another class of non-learning-based methods is simultaneously localization and mapping (SLAM), which deals not only with path-planning, as the aforementioned works, but also with obstacle detection and avoidance by constructing a map of the environment (\cite{Demim2018, Liu2021, Ren2022}). The drawback of SLAM-based methods is the fact that in large-scale environments, it will exhibit degraded efficiency since building a map of the whole environment is practically infeasible. \\
Urban environments can be described as large scale, complex geometries, with buildings representing dense obstacles and flow-fields characterized by turbulent flows. These features describe a problem with many uncertainty sources, to which also the characteristics of the UAVs have to be added, such as sensors noise for navigation, calibration and control errors. This leads to the necessity of using learning-based methods, which can mostly overcome and handle these critical aspects of the task of navigation and obstacle avoidance. Supervised deep Learning (DL)-based methods have received significant attention in recent years. DL can significantly enhance obstacle avoidance and path planning for UAVs by leveraging neural networks to process and interpret vast amounts of sensory data, such as images from cameras or signals from LiDAR (\cite{Tang2024, Osco2021, roghair2021}). This approach allows UAVs to detect and navigate around obstacles more efficiently by recognizing patterns and predicting potential collisions in real time. Supervised DL methods are extremely efficient for environments exhibiting small variations, as they are based on labels and are sensitive to environmental changes. These features make them suitable for closed environments, but less reliable for urban environments, where the conditions typically change very rapidly. This brings the necessity to develop reinforcement-learning (RL) methods, which are unsupervised, for understanding and automating decision making processes, in which the agent learns based on the given goals (\cite{Sutton2018}). \\
The optimal policy is obtained by learning how to map states to actions, and the agent learns through a trial-and-error procedure to get the actions which will ideally yield the quantitative highest reward and qualitatively best result depending on the task it has to execute. With respect to other methodologies, RL makes an agent learn by letting it directly interact with the environment. Deep reinforcement learning (DRL) combines the advantages of both DL and RL. In particular, DRL leverages the neural-network (NN) architectures of DL to approximate the value functions or policies used in RL in order to solve problems where the action space, the observation space or both are high dimensional (\cite{MATSUO2022}, \cite{Rao2024}). Moreover, DRL agents are able to generalize from raw input data by sensors or images to learn representations that capture the underlying structure of the environment, leading to more robust decision-making (\cite{Yang2018}, \cite{zhang2021}). Observations are directly mapped into actions to be taken by the agent, and this results in an end-to-end decision-making strategy which drastically reduces uncertainties due to sensor noise or low-quality inputs. The model of the environment is important to correctly guide the agent towards the goal but the algorithm does not strictly depend on it(\cite{farsang2021}, \cite{Ocana2023}). Indeed, since DRL aims to map the optimal relations between observations and actions, even if the environment changes the agent can still take the suitable actions related to the received observations. Another advantage with respect to supervised DL is that DRL does not need labels because the agent directly interacts with the environment and generates the reward signal used for learning on the fly. \\
Machine learning (ML) has experienced a rapid development in the last years and has transformed the state-of-the-art capabilities for many tasks in engineering and computer science. In particular, it has been exploited to enhance fluid-flow simulations for a variety of applications, from turbulence modelling to development of boundary conditions (\cite{Vinuesa2022},\cite{Vinuesa2022a}). In this context, DRL showed to be particularly suitable to face non-linear and high dimensional problems, such as turbulence and flow control(\cite{Vignon2023}).  \\
\noindent Active flow control is an extremely interesting topic in the field of DRL applications in fluid mechanics. \cite{Rabault_Kuchta_Jensen_Réglade_Cerardi_2019} applied DRL to a two-dimensional (2D) simulation of the flow around a cylinder to learn an active control strategy from varying mass flow rates of two jets on the sides of said cylinder, achieving a considerable drag reduction. This approach has been extended to a multi-environment configuration, which considerably speeded up the execution by adapting the DRL algorithm for parallelization (\cite{Rabault2019}). The active-flow-control problem around a cylinder has been extended to three dimensions (3D) by Suarez et al. (\cite{suárez2023activeflowcontrolthreedimensional}, \cite{suárez2024activeflowcontroldrag}), using a Multi-Agent Reinforcement Learning (MARL) approach coupled with a CFD solver, which led to a considerable drag reduction after applying the DRL control on three different configurations. MARL has been also successfully applied to a 2D Rayleigh--Bérnard convection problem, allowing to control this multiple-input multiple-output problem (\cite{Vignon2023a}), and to drag reduction in fully developed turbulent channels (\cite{Guastoni2023, Hasegawa_2023}). \\
These successful applications of DRL in flow control have led to its application in a variety of engineering tasks. Among them, path planning, trajectory optimization and obstacle avoidance using DRL received a massive interest in the last years, dealing with different models. \cite{Gunnarson} applied a DRL algorithm based in Remember-and-Forget-Experience-Replay to steer a fixed-speed swimmer through an unsteady 2D flow field. A modified twin delayed DDPG (TD3) model was used to execute a navigation task in multi-obstacle environments with moving obstacles (\cite{ZHANG2022}). Here, Zhang et al. wanted to predict the impact of the environment on the UAV and the change of observations was added in the actor-critic (AC) network. Then, a two-stream AC network structure was proposed to extract features of the observations provided in the environment. Another approach to the same multi-obstacle problem uses a modified recurrent deterministic policy gradient (RDPG) algorithm, named Fast-RDPG, in which the parameters of the policy are updated step by step, without the necessity to wait until the end of an episode to update them (\cite{Wang2019}). A variation of double deep Q-Network (DDQN), the Autonomous Navigation and Obstacle Avoidance (ANOA) algorithm, was developed by Wu et al.(\cite{Wu2020}). Here, the network was divided into two parts, one outputting a state-value function, the other outputting an advantage-value function, which ensures the extra-reward value of choosing an action rather than another. The action-advantage value is independent of state and environment noise, which are instead taken into account in the state-value function. Wang et al. developed an algorithm for military applications, in which the main focus was collision avoidance by including a Faster Region-based Convolutional Neural Networks (R-CNN) model and a Data Deposit Mechanism to extract information about the obstacles from images, based on a Deep Q-Network (DQN) algorithm (\cite{WANG2024}). Jin et al. proposed a multi-input attention prioritized deep deterministic policy gradient algorithm (MAPDDPG), which introduces an attention mechanism to help the UAV focus on environmental information relevant for the navigation task (\cite{Jin2021}). Despite the differences and enhancements tailored to each of these algorithms, they are all off-policy learning algorithms and use target networks, meaning that they learn from experiences collected using a different policy than the one is getting optimized. The experiences are stored in a replay buffer to break the correlation between consecutive experiences and stabilize training. Target networks are used to stabilize the training and are periodically updated with the weights of the main networks to reduce the variance of the updates. \\
A more robust algorithm is Proximal Policy Optimization (PPO). This a policy gradient method, primarily designed to balance exploration and exploitation while keeping a stable and efficient training. PPO is often considered more robust than TD3, DDPG and DQN and their variants because PPO uses a clipped surrogate objective function, ensuring that the policy updates are not too drastic. By clipping the probability ratios, PPO limits the change in the policy at each update step, avoiding large deviations that could make the training unstable. PPO is an on-policy algorithm. The policy is updated based on data collected from the current policy, making the updates more stable because they are based on the most recent interactions with the environment. In addition, PPO adjusts the step size dynamically based on the performance of the policy and it features a reduced sensitivity to hyperparameter settings. The drawback is that it may require more samples overall, compared to an off-policy method. \\ 
When PPO is combined with recurrent neural networks (RNNs), it becomes highly efficient for many engineering problems, in particular in trajectory optimization. Federici et al.(\cite{Federici2023}) used this architecture with long short-term me\-mo\-ry (LSTM) cells to build a meta-RL algorithm to achieve autonomous waypoint guidance for a six-rotor UAV in Mars' atmosphere, showing a substantial improvement with respect to the simple PPO algorithm. Hu et al. (\cite{Hu2024}) used a PPO+RNN architecture to design an escape flight vehicle against multiple pursuit flight vehicles, demonstrating that the use of RNN enhances the capability of PPO to train the agent to accomplish the given task. \\
In this work, we aim to develop a method for trajectory optimization and obstacle avoidance for a UAV in a 2D domain characterized by a complex flow field. The approach to this problem can be described as a Zermelo's problem (\cite{Zermelo1931berDN}), but adding obstacles in the field. A formulation of the solution of the Zermelo's problem by means of RL was given by Biferale et al. (\cite{Biferale2019}), where a 2D velocity flow field derived from numerical simulations was given, comparing the efficiency of the RL approach with that of a classic optimal navigation method. This configuration was used as validation step of the methods used here, with substantial modifications in the algorithm. The problem was apporached with PPO and the complexity was increased by having random starting and target areas. PPO resulted in a more stable training, reaching a reward at convergence slightly higher than using an AC algorithm.\\
The problem stated here can be considered as a partially observable markov decision process (POMDP). POMDP is substantially different from classical markov decision process (MDP), where the main feature of MDPs is that the future state depends only on the current state and action, and not on the sequence of events that preceded it. POMDPs extend MDPs to cases where the agent cannot directly observe the true state of the environment. Instead, it receives observations that provide partial information about the state. The agent does not have direct access to the full state of the environment but must rely solely on observations that provide incomplete information. In this context, we have only a partial representation of the environment and the state of the UAV, since it would be too complex and computationally unfeasible to map all the points and variables of the domain, considering that we are using results from a high-fidelity simulation of a turbulent flow field. This makes the problem even more challenging, since the agent has to deal with uncertainties not only of the environment and the turbulent flow field, but also of the observability of the state itself. \\
To the best of the author's knowledge, there are no existing studies that specifically address UAV trajectory optimization using PPO and LSTM in a turbulent flow-field generated from high-fidelity numerical simulations, particularly in scenarios involving obstacle avoidance. While LSTM-based models have been previously used in reduced-order modeling (ROM) of turbulent flows, capturing temporal dynamics from direct numerical simulations (DNS) data, these models have primarily focused on learning and predicting flow behavior without integrating them into UAV navigation tasks (\cite{mohan2018deeplearningbasedapproach}). Additionally, recent work on UAV obstacle avoidance and deep reinforcement learning typically utilizes partially observable environments with LSTMs, but these efforts have not yet extended to scenarios involving complex CFD-generated turbulent flow fields \cite{singla2018memorybaseddeepreinforcementlearning}. Current works in UAV navigation that use PPO generally focus on simplified or structured environments, such as urban spaces involving static obstacles  or moving entities, without incorporating the complexities of turbulent flow-fields, or multi-agent obstacle avoidance systems (\cite{HuDong2023}, \cite{Hu2022}). The combination of PPO with LSTM networks has been applied to multi-agent cooperative systems and collision avoidance, but these studies do not incorporate complex turbulent environments \cite{rao2023integratedrealtimeuavtrajectory}. \\
The method proposed in this work, which integrates snapshots of DNS-generated flow fields into a reinforcement learning framework, addresses this gap. It allows for the precomputed turbulence data to inform the UAV’s navigation decisions, offering a higher-fidelity representation of real-world fluid dynamics compared to the methods used in these prior studies. This approach enhances the realism and complexity of the environment without the computational burden of real-time CFD interaction, setting it apart from previous works.
The present work is structured as follows: In Section \ref{setup}, the difference between MDP and POMDP is explained; the problem addressed here is also described, including the environment and the architecture of the chosen algorithm. In Section \ref{results}, the results of the application of the algorithm to the problem are shown and compared with the results of PPO and TD3 algorithms. In Section \ref{Conclusions}, the main results are summarized and future work is proposed. 

\section{Problem statement}
\label{setup}

This work is focused on the navigation of a UAV in a 2D slice of a 3D turbulent flow-field, with obstacles representing buildings which the UAV has to avoid. The main goals are obstacle avoidance (spatial problem) and trajectory optimization (temporal problem), in a way that the UAV can reach a target finding the safest and quickest path. The task is designed to be addressed using DRL, in particular using a PPO+LSTM architecture.

\subsection{MDPs and POMDPs}
The main features of an MDP are a state space $S$, and initial state space $S_0$ with an initial state distribution $p(s_0)$, an action state space $A$, a state transition probability distribution $p(s_{t+1}|s_t,a_t)$ which satisfies the Markov property, and a reward function $r(s_t,a_t):S \times A \rightarrow R$. The reward function provides the feedback of the environments when executing the action $a_t$ in a state $s_t$. State and action spaces can be continuous or discrete. These give a first indication about how to build the architecture and the network associated with the DRL algorithm. RL is generally used to solve a MDP. An optimal policy is learned through a trial-and-error process, and the policy could be stochastic or deterministic. A stochastic policy $a \sim  \pi(\cdot|s):S\rightarrow P(A)$ returns the probability density of available state and action pairs ($s,a$), where $P(A)$ is the set of probability measures on $A$. Note that a deterministic policy $\mu(S):S\rightarrow A$ only projects states into actions. \\
A POMDP is characterized by the fact that the agent cannot directly observe the state $s_t$, but receives a set of observations $o_t$ with a distribution $p(o_t|s_t)$. The sequence of observations do not satisfy the Markov property, since $p(o_{t+1}|a_t, o_t, a_{t-1}, o_{t-1}, ..., o_0 )\neq p(o_{t+1}|o_t,a_t)$. Consequently, the agent has to infer the current state $s_t$ based on the history of trajectories.\\
In this work, we exploit in particular the ability of LSTMs to capture temporal dependencies in a POMDP, in combination with PPO, in an environment characterized by a partial observability, in order to design an algorithm powerful enough to be prone to extension to a 3D problem and to even more uncertain environments. The detailed description of policy-gradient methods, PPO, LSTMs and their combination is described in detail in the Appendix.\\ 

\subsection{Fluid-flow data}
\begin{figure}[h]
    \centering
    \includegraphics[width=\linewidth]{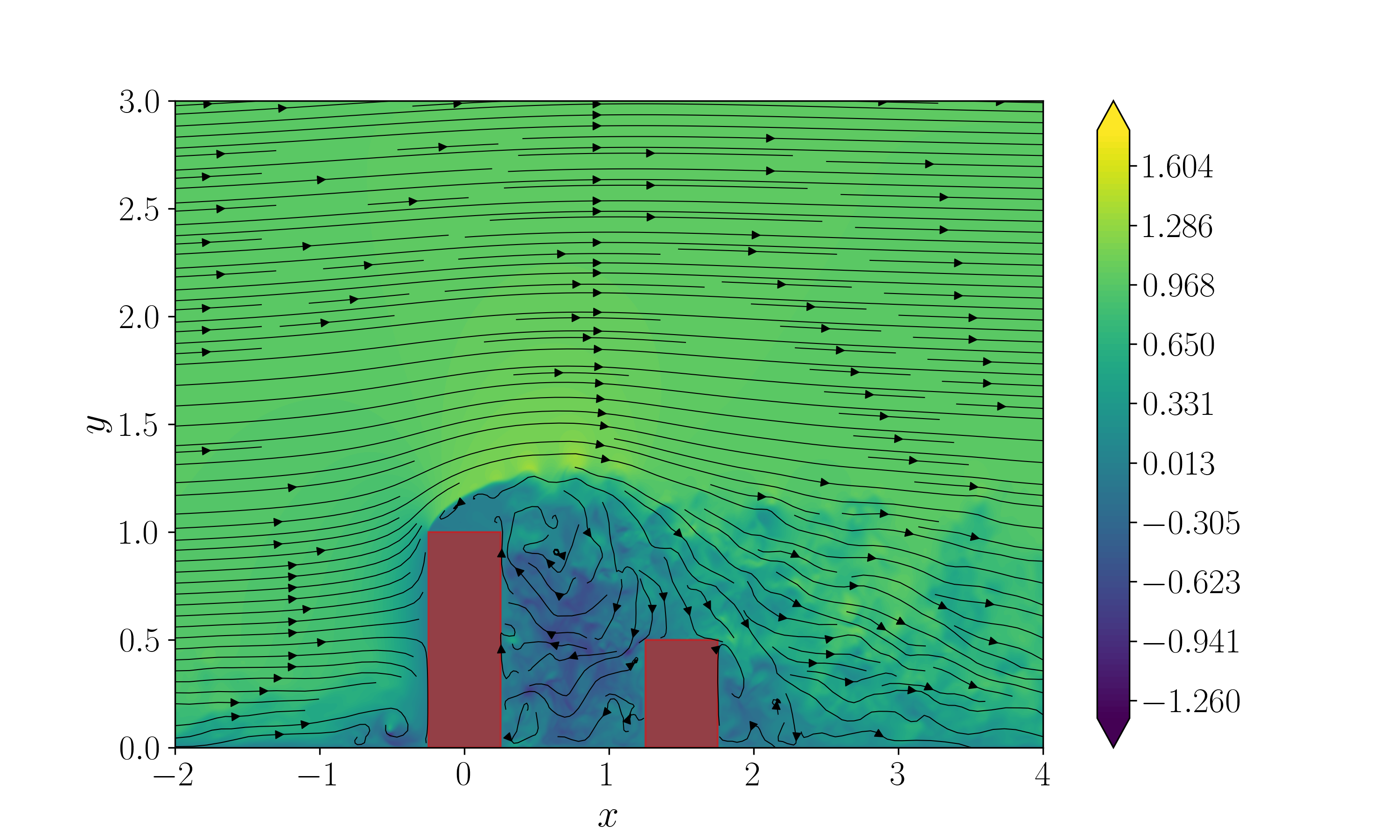}
    \caption{A snapshot showing the streamwise velocity $U$ of a 2D slice of the instantaneous flow-field with obstacles. Obstacles are highlighted as red rectangles. }
    \label{flowfield}
\end{figure}

The flow field is represented by a 2D data set extracted from 3D high-fidelity simulations performed with the spectral-elements code Nek5000 following the same numerical setup as in Atzori et al. \cite{Atzori2023} and Zampino et al (\cite{zampino2024}). The flow field is extracted at the centerplane plane $z=0$ (where $x$, $y$ and $z$ are the streamwise, vertical and spanwise coordinates, respectively). The domain coordinates\footnote{All the distances are scaled by the obstacle height $h$.} are $x\in[-2.0,4.0]$, $y\in[0,3.0]$. Obstacles coordinates are $x_{o1}$$\in[-0.25,0.25]$, $y_{o1}$$\in[0.0,1.0]$, $x_{o2}$$\in[1.25,1.75]$, $y_{o2}$$\in[0.0,0.5]$. The dataset used the training consists in a set of 300 snapshots separated by 0.08750 time units, with time span of the dataset of 26.25 time units\footnote{Time is normalized with $h$ and the freestream velocity $U_{\infty}$}. Figure \ref{flowfield} shows the streamwise velocity $U$ of a 2D slice of an instantaneous flow field. The flow field shows both recirculation zones and areas where $U$ is much greater than the surroundings, indicating the paths which could be more challenging for the safe navigation of the UAV.

\subsection{UAV dynamics}
UAV navigation is typically described with a set of non-linear differential equations in a three-dimensional space. Since in the present work the flow field is two dimensional, the UAV is modelled as a mass point and the set of non-linear equations is reduced to a 2D space. The variables describing the UAV motion here are: position vector ($\boldsymbol{x}$), velocity vector($\boldsymbol{v}_{\textrm{global}}$), orientation (heading angle, $\theta$) and angular velocity ($\omega$). The dynamics and state of the UAV are significantly influenced by the underlying flow field.
The equations of motion take into account the presence of the surrounding flow field with velocity components \(u_{\text{flow}}\) and \(v_{\text{flow}}\), and are given by:

\begin{equation}
    \begin{cases}
    \frac{\mathrm{d}\boldsymbol{x}}{\mathrm{dt}} &= \boldsymbol{v}_{\textrm{global}} + \boldsymbol{v}_{\textrm{flow}}, \\
    \frac{\boldsymbol{v}_{\text{global}}}{\mathrm{dt}} &= a \begin{pmatrix} \cos(\theta) \\ \sin(\theta) \end{pmatrix}, \\
    \frac{\mathrm{d}\theta}{\mathrm{dt}} &= \omega, \\
    \frac{\mathrm{d}\omega}{\mathrm{dt}} &= \Dot{\omega},   
    \end{cases}\
\end{equation}

\noindent where \(\vec{v}_{\text{flow}} = \begin{pmatrix} u_{\text{flow}} \\ v_{\text{flow}} \end{pmatrix}\) represents the velocity of the flow field at the UAV's position, while $a$ and $\Dot{\omega}$ represent the linear and angular accelerations, respectively. $a$ and $\Dot{\omega}$ are also the controls which represent the actions taken bu the UAV when interacting with the two dimensional flow-field. \\
To solve the equations of motion, the fourth-order Runge--Kutta method is used, yielding the state of the UAV at the next time step.

\subsection{Environment}
\label{env}
\begin{figure}
\centering
\includegraphics[width=0.5\linewidth]{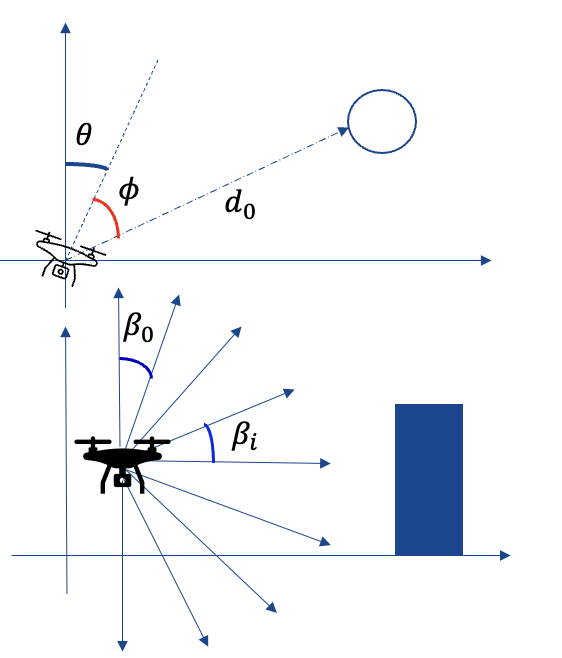}
\caption{Schematic representation of the observation space described in Eq.~(\ref{obsspace}).}
\label{observationspace}
\end{figure}

Both observation and action spaces are continuous. The observation space is desgined as follows:
\begin{equation}
    o = \left\{\theta,\phi,d_{0}\right\}+\left\{\beta_{i}\right\},
    \label{obsspace}
\end{equation}
where $\theta$ is the heading angle of the UAV, $\phi$ is the relative angle of the UAV with respect to the target, $d_0$ is the distance between of the UAV and the target and $\left\{\beta_{i}\right\}$ are the are the angles associated with the sensors for obstacle detection, with $i\in[0,8]$ spanning the angles between $-\pi$ and $\pi$. Figure \ref{observationspace} shows graphically the definition of the observation space.

\begin{figure}
\centering
\includegraphics[width=0.6\linewidth]{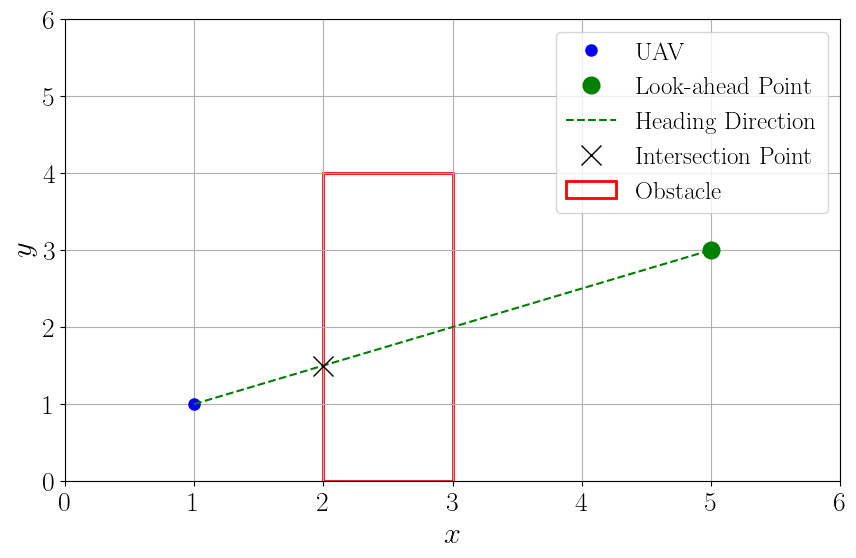}
\caption{Sketch of the obstacle detection procedure.}
\label{obstacledetection}
\end{figure}

The action space is designed to include the linear and angular accelerations of the UAV, where $a\in[-3.0,3.0]$ and $\dot{\omega}\in[-\pi/4,\pi/4]$, while setting constraints to the velocity, which is bounded in magnitude, $v_{\textrm{UAV}}\in[-1.4v_{\textrm{max}},1.4v_{\textrm{max}}]$, where $v_{\textrm{max}}$ is the magnitude of the maximum velocity of the flow field across all the frames used by the algorithm. \\
From the formulation of the observation space, it can be observed that obstacle detection is achieved by providing to the agent a set of directions, since in UAVs the relation with the surroundings is typically given by images from cameras, radar signals or range finders. In this work, we take as inputs for the observation space the angles which represent the orientation of rays sent by range finders mounted on the UAV. Obstacle detection is achieved by implementing a ray-tracing technique (\cite{TONTI2021}). First of all, the UAV has to check for free space in its perspective. The input is the position of the UAV and the output is a boolean variable which indicates whether the path is free from obstacles or not. Then, if the obstacle is present, the intersection with the traced rays is computed. First, the direction of the ray is calculated, based on the ray origin and final point, as well as the coordinates of the obstacles. Then, it is verified whether parallel directions to the obstacles are present. If the detected directions are not parallel to the obstacles, the intersection point between the ray and the obstacle is calculated and the distance to the intersection is returned. Figure \ref{obstacledetection} sketches the process.

The starting and target areas are chosen randomly before the first obstacle and after the second obstacle, respectively.
The agent is allowed to take a maximum of 80 steps in the environment for each episode. The starting frame of the simulation is random, meaning that the navigation task does not always start from the same simulation data, but changes randomly for every episode, so that the initial conditions of the flow field themselves exhibit uncertainties. \\
The UAV state is described as follows:
\begin{equation}
    s=\left\{\boldsymbol{x},\theta,\boldsymbol{v},\omega\right\},
\end{equation}
where $\boldsymbol{x}$ is the position vector, $\theta$ is the heading angle, $\boldsymbol{v}$ is the velocity vector and $\omega$ is the angular velocity.\\
The state of the UAV is inferred from the observations, which are given as input to the NN and described in Equation (\ref{observationspace}).
The environment has been implemented using the format of OpenAI Gymnasium (\cite{towers_gymnasium_2023}).

\subsection{Reward function}
\label{reward_sec}
As mentioned in Section \ref{introduction}, DRL is a process that encourages learning by trial and error and this process is triggered by a reward which is given to the agent when it takes the right actions to complete the assigned task. The structure of the reward is crucial because this guides the agent towards a more effective learning, so this component of the algorithm has to be carefully designed and tuned for a specific task. The reward structure of this work is inspired by other obstacle-avoidance and navigation problems (\cite{ZHANG2022}, \cite{Wang2019}, \cite{Jin2021}). \\
The reward structure is designed to guide the UAV towards the target while minimizing collisions with obstacles, reducing energy consumption and preventing leaving the designated operational bounds. The reward function is constructed from several components, each one addressing a different aspect of the UAV's performance. The final reward is a sum of the following components:  
\begin{itemize}
    \item Transition reward $r_{trans}$: it describes the progression of the UAV towards the target, and is defined as:
    \begin{equation}
        r_{\mathrm{trans}} = \sigma d_{\mathrm{dist}},
    \end{equation}
    where $\sigma\in\mathbb{R}$ is a scaling factor to weight the influence of this term in the total reward, and $d_{\textrm{dist}}$ is the reduced distance from the starting point to the target:
    \begin{equation}
        d_{\mathrm{dist}} = \left\Vert\ x_{t-1} - x_{\mathrm{target}}\right\Vert - \left\Vert\ x_{t}-x_{\mathrm{target}}\right\Vert,
    \end{equation}
    where $x_{t-1}$ is the position of the UAV at the previous time step, $x_{\textrm{target}}$ is the position of the target point and $x_{t}$ is the current position of the UAV;
    \item Obstacle penalty $r_{\textrm{obs}}$: it detects how close the UAV is to the obstacle. The closest the UAV is to the obstacle, the larger the value of the penalty, described here as:
    \begin{equation}
       r_{\textrm{obs}} = -\alpha e ^{-\psi d_{\textrm{min}}},
    \end{equation}
where $\alpha$, $\psi$ $\in \mathbb{R}$ are constants, and $d_{\textrm{min}}$ = $\textrm{min} \left\{d_{1},..,d_{n}\right\}$, with $n\in$(0,8) being the minimum of the distances between the UAV and the obstacles provided by the sensor measurement.
    \item Free-space reward $r_{\textrm{free}}$: it is the reward assigned for navigating in a direction free of obstacles:
    \begin{equation}
    r_{\textrm{free}} = 
    \begin{cases} 
    r_{\textrm{free}} & \text{if free space ahead is detected} \\
    0 & \text{otherwise},
    \end{cases}    
    \end{equation}
    where $r_{\textrm{free}}\in\mathbb{R}$ is a constant.
    \item Best-direction free-space reward $r_{\textrm{best}}$: if the path ahead is not free of obstacles, the agent chooses an alternative direction:
         \begin{equation}
    r_{\textrm{best}} = 
    \begin{cases} 
    \zeta\beta_{\textrm{best}} & \text{if no free space ahead} \\
    0 & \text{otherwise},
    \end{cases}      
    \end{equation}
    where $\zeta\in\mathbb{R}$ is a constant and $\beta_{\textrm{best}}$ is the angle which returns a direction representing the maximum distance from the obstacle in the set detected by the sensors;
    \item Step penalty $r_{\textrm{step}}$: it is the penalty given for each step taken by the agent, which encourages the UAV to find the shortest path to the target, $r_{\textrm{step}}\in\mathbb{R}$;
    \item Energy penalty $r_{\textrm{energy}}$: it accounts for the UAV's propulsion energy consumption. It is calculated as:
    \begin{equation}
    r_{\textrm{energy}} = -\kappa|| \mathbf{v_{\textrm{prop}}} ||,     
    \end{equation}
    where $\mathbf{v_{\textrm{prop}}}$ is the propulsion velocity, defined as the difference between the UAV's current velocity and the flow velocity. This penalty discourages excessive energy use by reducing the reward proportionally to the effort required to move relative to the surrounding flow.
\end{itemize}
The total reward is then given by:
\begin{equation}
    R_{\textrm{tot}} = \sum_{i=0}^{m}r_{\textrm{trans}_{i}}+r_{\textrm{obs}_{i}}+r_{\textrm{free}_{i}}+r_{\textrm{best}_{i}}+r_{\textrm{step}_{i}}+r_{\textrm{energy}_{i}},
\end{equation}
where $m\in\mathbb{N}$ is the number of steps taken by the agent in the environment.
At the end of each episode of $m$ steps, additional terms are considered. If the agent reaches the target it gets an extra reward, whereas if it hits an obstacle it receives an extra penalty, as well as if it hits the bounds of the domain.

\subsection{PPO+LSTM}
\label{algo_descr}
\begin{figure}
\centering
\includegraphics[width=0.8\linewidth]{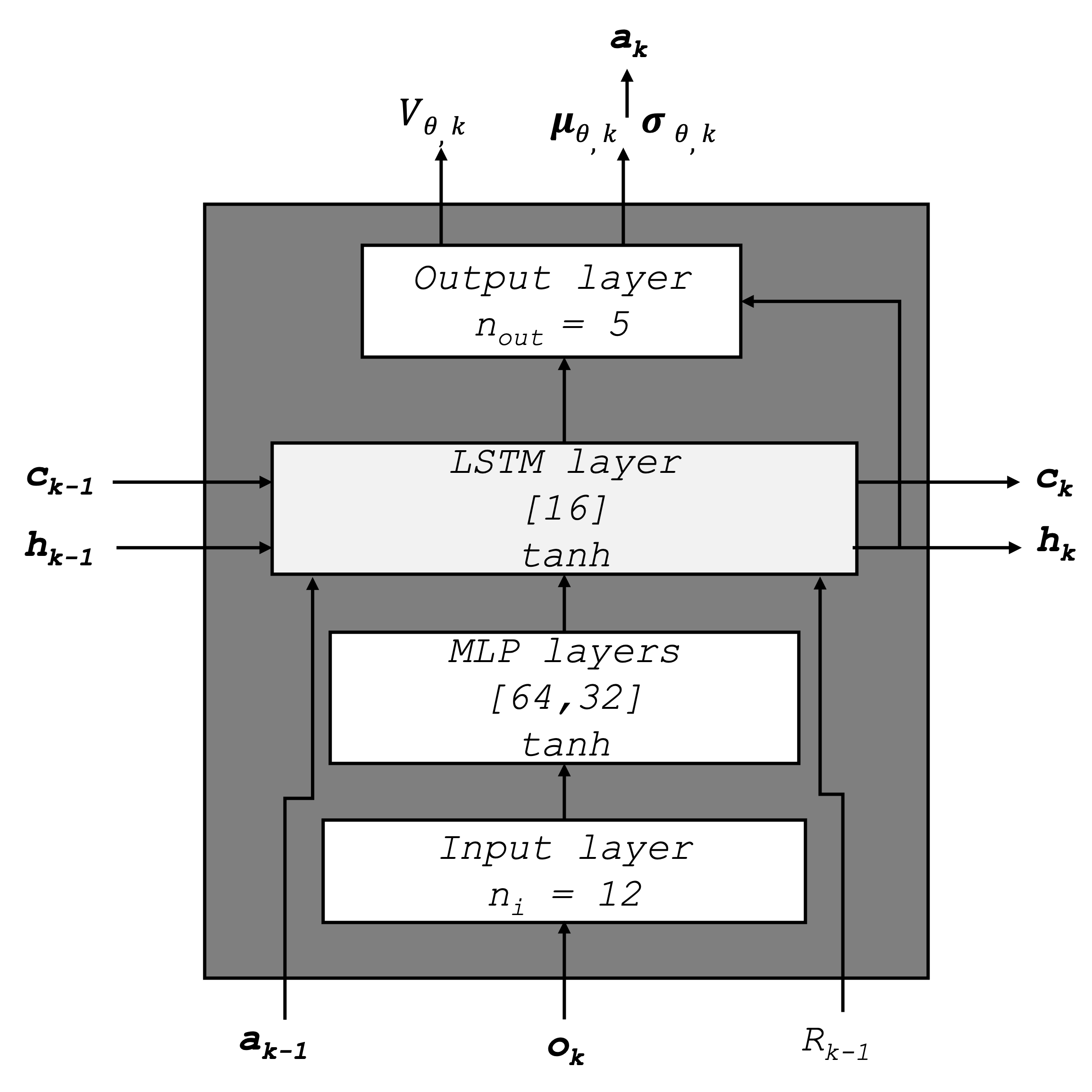}
\caption{Architecture of the PPO+LSTM network. $n_{i}$ denotes the number of observations fed into the NN, equivalent to the dimensions of the observation space. The number of outputs $n_{\textrm{out}}$ corresponds to the mean and standard deviation of each of the actions ($\boldsymbol{\mu}_{\theta, k}$ and $\boldsymbol{\sigma}_{\theta, k}$), which are two, and the value function, estimating the expected return from the current state ($V_{\theta, k}$).}
\label{ppo+lstm}
\end{figure}

The algorithm used in this work has been completely developed in the context of this work, and is written in PyTorch (\cite{pytorch}).
Figure \ref{ppo+lstm} shows a sketch of the network structure.

The first layer takes as input the vector of observations ($\boldsymbol{o_\textrm{k}}$) described in Section \ref{env} and feeds the raw observations into the first hidden layer of the neural network. The first hidden layer is a fully connected layer with 64 neurons and a hyperbolic tangent (tanh) activation function, which allows to capture the non-linear relationships in the data. The second hidden layer is a again a fully connected layer, with a reduced size of 32 neurons, again with a tanh activation function. The LSTM layer follows the two dense layers, with 16 neurons. It processes the temporal features and models the temporal dynamics of the problem. The LSTM layer takes as additional inputs also the last reward ($R_{k-1}$) and last actions ($\boldsymbol{a_{\textrm{k-1}}}$), thus benefiting from the memory of past experiences. Finally, the network outputs are divided into two distinct streams. The first stream produces parameters for each action in the action space, specifically the mean ($\mu$) and standard deviation ($\sigma$), which are used for sampling actions in a stochastic policy. The second stream outputs a single scalar value representing the value function, which estimates the expected return from the current state. The output layer uses a linear activation function, appropriate for both continuous action distribution parameters and value estimation. Note that $\boldsymbol{c}$ and $\boldsymbol{h}$ represent the parameters of the hidden state which are fed and then given as output of the LSTM layer. \\
The hyperparameters of the algorithm were tuned by a trial-and-error procedure and are summarized in Table \ref{tab:ppo_hyperparameters}. The Adam optimizer (\cite{Adam}) is chosen because of its robustness and adaptive property.

\begin{table}[h!]
\centering
\caption{Summary of hyperparameters.}
\label{tab:ppo_hyperparameters}
\begin{tabularx}{\linewidth}{|l|c|X|}
\hline
\textbf{Hyperparameter}      & \textbf{Value} & \textbf{Description}                                \\ \hline
Clip ratio                   & 0.2            & Controls the range of policy updates (clipping)     \\ \hline
Stochastic gradient descent (SGD) steps & 30 & Number of SGD steps per epoch                       \\ \hline
Discount factor ($\gamma$)   & 0.99           & Discount factor for future rewards                  \\ \hline
GAE lambda ($\lambda$)       & 0.95           & Balances bias and variance in advantage estimation  \\ \hline
Batch size                   & 256            & Number of samples per batch                         \\ \hline
Mini-batch size              & 128            & Number of samples per mini-batch                    \\ \hline
Policy update epochs         & 4              & Number of epochs to update the policy               \\ \hline
Learning rate                & $10^{-4}$      & Initial learning rate of Adam optimizer             \\ \hline
Evaluation interval          & 5              & Frequency of model evaluation during training       \\ \hline
\end{tabularx}
\end{table}

\begin{figure}[htbp]
    \centering
    \begin{subfigure}[b]{0.6\textwidth}
        \centering
        \includegraphics[width=\textwidth]{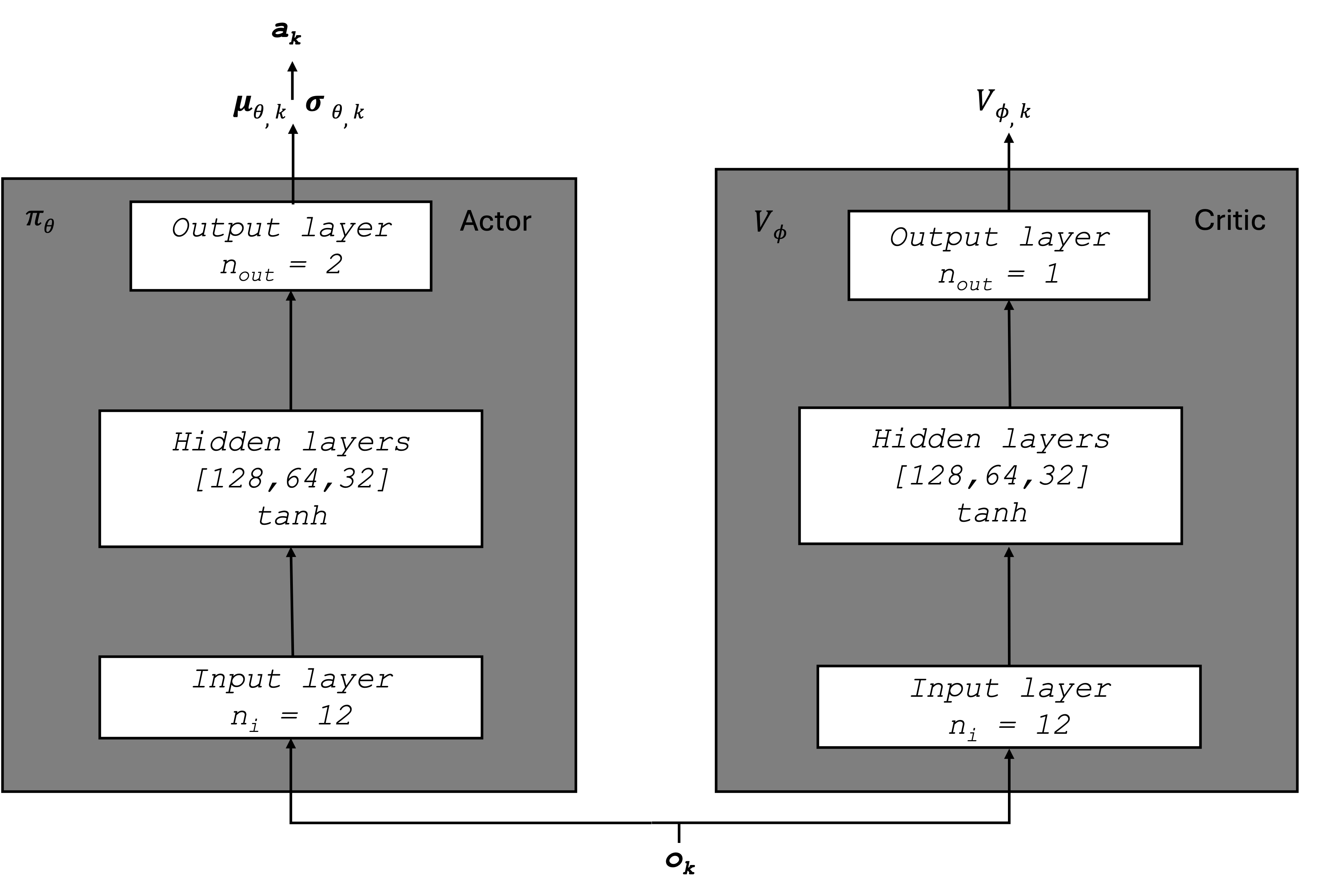}
        \caption{Schematic of PPO}
        \label{fig:subpol1}
    \end{subfigure}
    \vfill
    \begin{subfigure}[b]{0.6\textwidth}
        \centering
        \includegraphics[width=\textwidth]{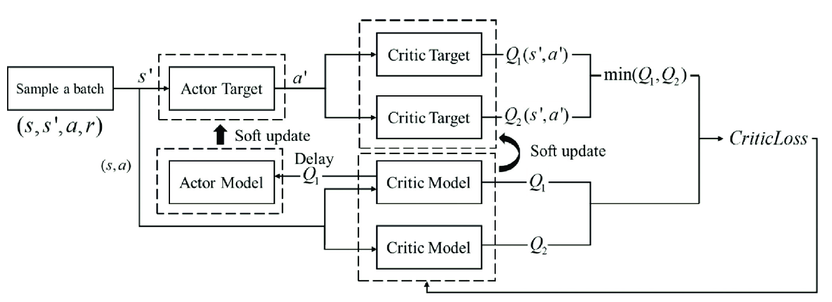}
        \caption{General structure of TD3}
        \label{fig:subpol2}
    \end{subfigure}
    \caption{Schematic representations of the PPO and TD3 architectures. The sketch of the TD3 network in panel (b) is extracted from \cite{TD3}.}
    \label{ppo+td3}
\end{figure}

The results obtained with a PPO+LSTM approach are compared with the results obtained applying a TD3 and a traditional PPO algorithm, also custom implemented, which are widely used for similar tasks in less complex environments. The structures of these networks are described in Figure \ref{ppo+td3}.

For Figure \ref{fig:subpol1}, the same nomenclature as for Figure \ref{ppo+lstm} applies. Figure \ref{fig:subpol2} describes the classical structure of a TD3 architecture. In our implementation, the actor network comprises 5 fully connected layers with 256, 128, 64, 32, and 2 nodes, respectively. Rectified-linear-unit (ReLU) activations are applied after each of the first four layers, with the final layer using a tanh activation scaled by the maximum action value. The critic network also has 5 fully connected layers with 256, 128, 64, 32, and 1 node. It takes both the state and action as input, concatenates them, and processes the combined input through ReLU activations after each layer to output a single Q value.
\noindent The learning rates for the actor and critic networks are set to $10^{-4}$ and $10^{-3}$, respectively, for both PPO and TD3. 

\section{Results and discussion}
\label{results}

\begin{figure}
\centering
\includegraphics[width=0.8\linewidth, scale=2.0]{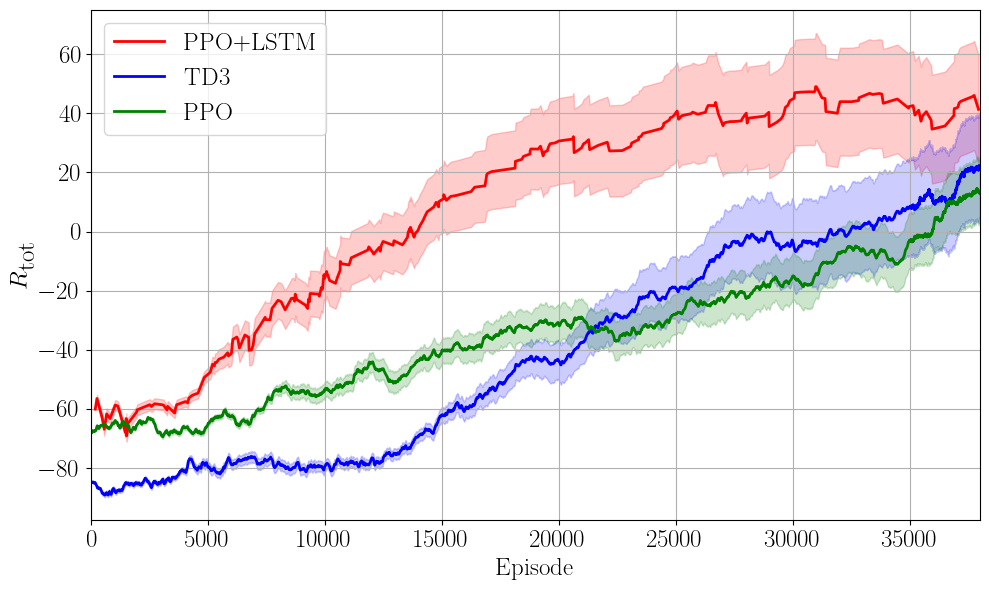}
\caption{Comparison of the total rewards against episode number for PPO+LSTM (red), TD3 (blue) and PPO (green)}
\label{rewards_comp}
\end{figure}

In this section, the main results obtained with the present methodology are presented. 
Figure \ref{rewards_comp} shows the comparison of the rewards of PPO+LSTM, the TD3 and the PPO algorithms. The reward is represented with an exponential moving average (EMA) for visualization purposes, together with the standard deviation of the reward. EMA is a technique used to smooth data by applying exponentially decreasing weights to past observations. This approach prioritizes recent data points, making the EMA more responsive to recent changes while still considering the entire history of the data. This exponentially smoothed value effectively captures underlying trends by reducing the impact of short-term fluctuations, making it ideal for visualizing noisy data in applications like model training. The EMA is calculated as follows:

\begin{equation}
\text{EMA}_t = \alpha\nu_{t} + (1 - \alpha)\text{EMA}_{t-1},
\end{equation}

\noindent where $\nu_{t}$ represents the value of the data at time $t$ and $\alpha\in$[0,1] is the smoothing factor, with values close to 0 corresponding to maximim smoothing and 1 corresponding to raw data.

\begin{figure}[htbp]
    \centering
    \begin{subfigure}[b]{0.49\textwidth}
        \centering
        \includegraphics[width=\textwidth]{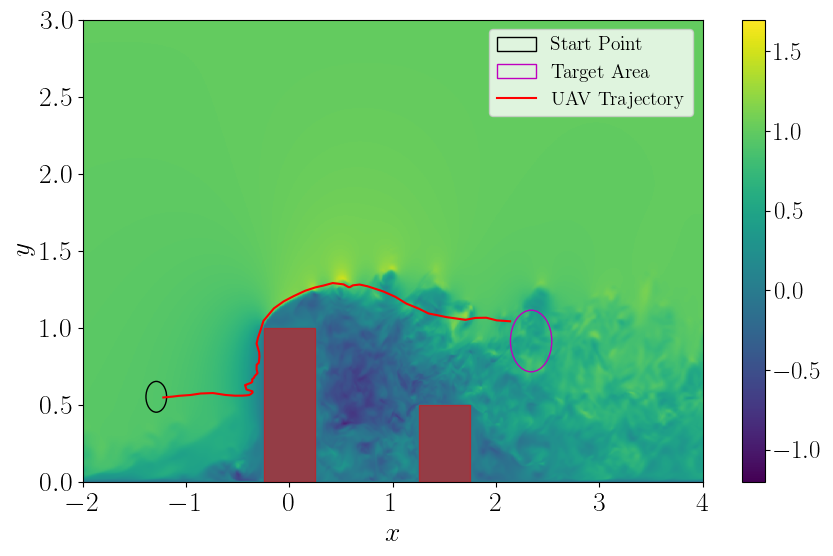}
        \caption{}
        \label{fig:sub1}
    \end{subfigure}
    \hfill
    \begin{subfigure}[b]{0.49\textwidth}
        \centering
        \includegraphics[width=\textwidth]{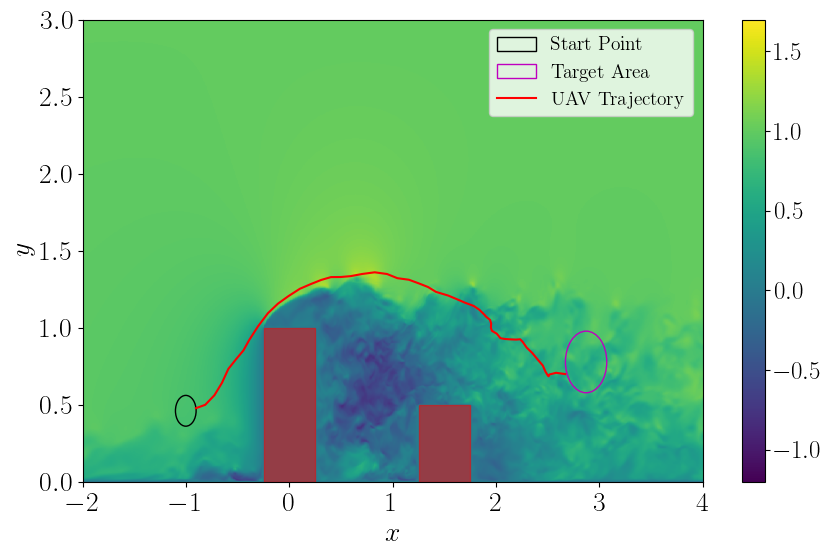}
        \caption{}
        \label{fig:sub2}
    \end{subfigure}
    \vskip\baselineskip
    \begin{subfigure}[b]{0.49\textwidth}
        \centering
        \includegraphics[width=\textwidth]{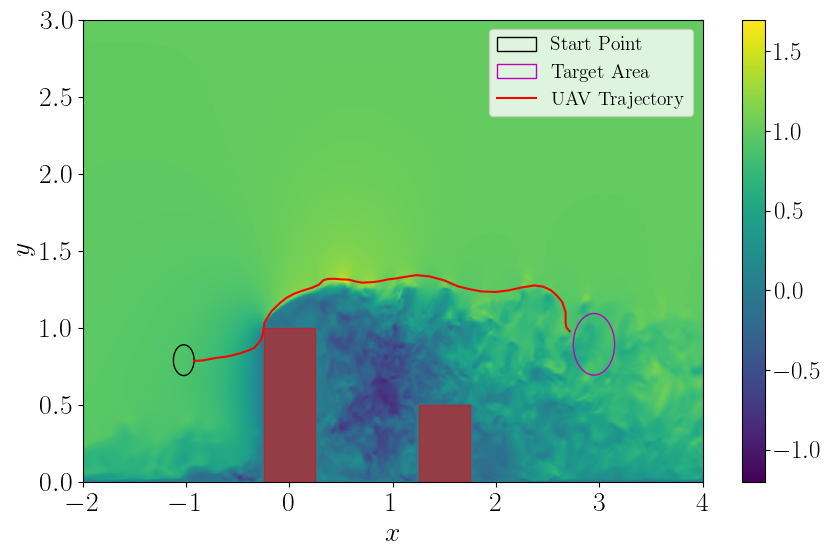}
        \caption{}
        \label{fig:sub3}
    \end{subfigure}
    \hfill
    \begin{subfigure}[b]{0.49\textwidth}
        \centering
        \includegraphics[width=\textwidth]{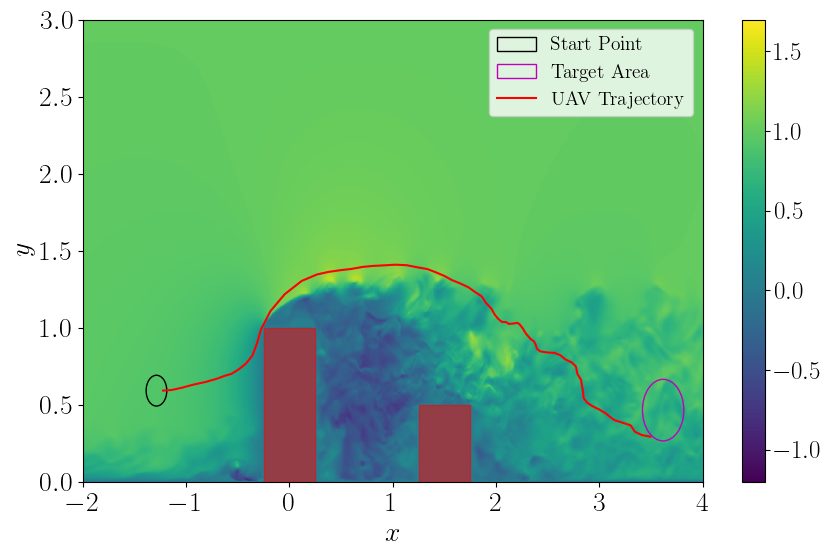}
        \caption{}
        \label{fig:sub4}
    \end{subfigure}
    \caption{Various trajectories of the UAV for different starting and target points produced during the evaluation phase of the final policy. The surrounding flow-field is displayed by the streamwise velocity.}
    \label{Trajectories}
\end{figure}

It can be observed that the PPO+LSTM combination outperforms  both the PPO and TD3 apporoaches. Not only the final value of the reward, but also the slope during training shows that convergence is much faster for PPO+LSTM with respect to the other two algorithms. This is due to the fact that the LSTM cells enhance memory of the recent events in the current episode trajectory and learning is much more effective than in architectures which do not have such memory capabilities. The other thing to take into account is the partial observability of the environment. In particular, the observation space does not include any information about the flow field, which is taken into account only when integrating the equations of the dynamics. The only information about the environment is given by the relative orientation to the target and the calculated distances from the obstacles. The state of the UAV is described by its position and orientation with respect to the target, its velocity and calculated distances from the obstacles, but the UAV state is not fully visible in the observations passed as an input to the NN. Based in the result, the PPO and TD3 algorithms without the contribution of RNNs are not suitable for POMDPs and problems where keeping track of the temporal sequence of the actions is crucial. A recent work by Wang et al. (\cite{Wang_2024}) developed a dynamic feature-based DRL (DF-DRL) for flow control which can provide a suitable alternative to the use of RNNs, which could be adapted and tested in trajectory-optimization problems. \\
\indent Several trajectories produced by the PPO+LSTM policy during evaluation are shown in Figure \ref{Trajectories}. 

\begin{figure}[htbp]
    \centering
    \begin{subfigure}[b]{0.8\textwidth}
        \centering
        \includegraphics[width=\textwidth]{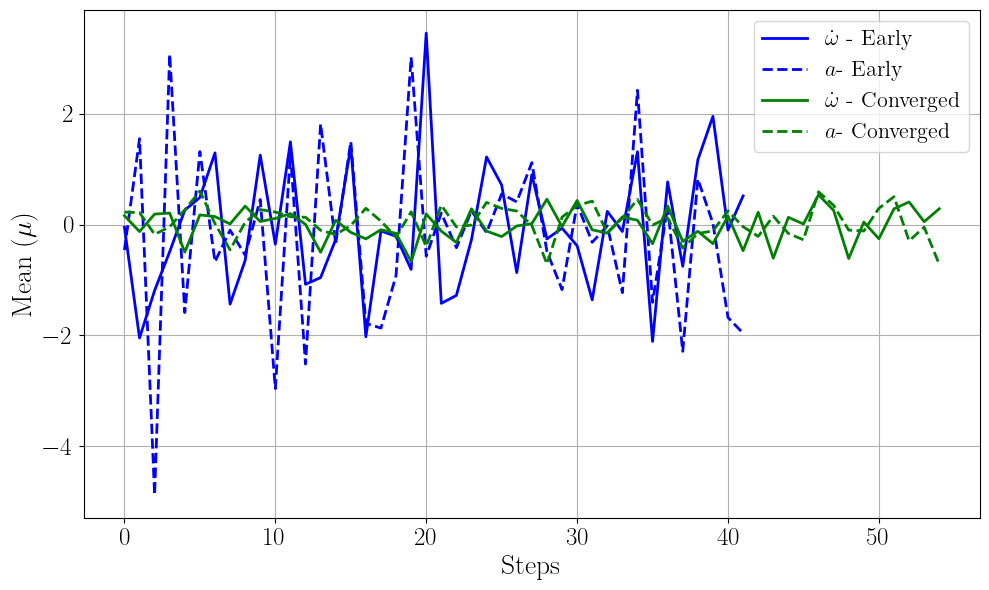}
        \caption{Evolution of $\mu$}
        \label{fig:mean}
    \end{subfigure}
    \vfill
    \begin{subfigure}[b]{0.8\textwidth}
        \centering
        \includegraphics[width=\textwidth]{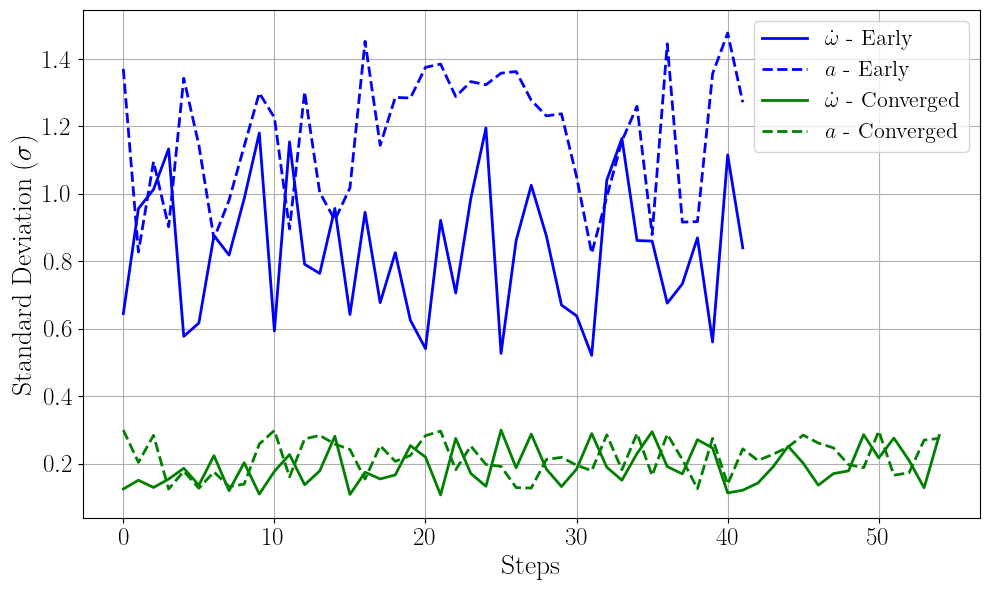}
        \caption{Evolution of $\sigma$}
        \label{fig:standard}
    \end{subfigure}
    \caption{Evolution of mean ($\mu$) (a) and standard deviation ($\sigma$) (b) of each action for each step in the environment during an episode in the early stages of training and at the converged policy.}
    \label{mean_and_stan}
\end{figure}

\noindent It can be observed that the mass point which represents the UAV not only avoids the obstacles, but also manages to properly exploit the flow-field regions where the velocity is higher, and avoids getting trapped in regions with high recirculation.
The success rate (SR) of the PPO+LSTM-trained policy reached 98.7\%, and the crash rate (CR) was 0.1\%. These results are significantly better than the ones obtained with PPO (SR = 75.6\%, CR=18.6\%) and TD3 (SR=77.4\% and CR=14.5\%), and highlight the importance of the memory cells, which are essential in navigation problems in complex environments. Moreover, the PPO+LSTM model requires fewer neurons to achieve much better performance than the other two networks, which are double the size. \\
\noindent The instantaneous mean ($\mu$) and standard deviation ($\sigma$) obtained from the last layer of the NN for each of the actions help us understand the agent's behavior and its exploration-exploitation trade-off during the learning process. Figure \ref{mean_and_stan} shows the evolution of $\mu$ and $\sigma$ fr each action at each step taken from the agent in the environment for two different episodes, one at early stages and one at the converged policy stage of the training.

\noindent The figure clearly shows that in the early stages of the training, the agent is still in exploration phase. This is visible from the blue lines in Figure \ref{fig:mean} and Figure \ref{fig:standard}, where the values of $\mu$ change significantly from step to step and from the fluctuations of the standard deviation $\sigma$ around high values, a fact that denotes that the policy is not converged.  On the other hand, at the final stages of the training (green lines of Figures \ref{fig:mean} and \ref{fig:standard}), both $\mu$ and $\sigma$ stabilize around lower values. The $\mu$ values indicate that the agent effectively manages to save energy when navigating through the flow, taking minimal corrections of the linear and angular accelerations and remaining essentially constant around a value; $\sigma$ in the same way fluctuates around values close to zero, indicating that the policy is essentially converged and that the exploration phase occurring in the early stages of the training has ended; thus the agent is exploiting the most effective actions learned during the training process. This is consistent with what we see with the trajectory visualizations in Figure \ref{Trajectories}, where the UAV effectively exploits the flow field to navigate through the environment following the vortical structures of the flow field to reach the target. 
\noindent Overall, these results provide insights into the agent's learned dynamics, highlighting the transition from a broad exploratory phase to a more focused exploitation of learned strategies. The increased concentration of actions in later episodes reflects the agent's improved understanding of the environment and its ability to identify and repeat actions that contribute to achieving its objectives.\\
With respect to other works (\cite{ZHANG2022,WANG2024,Wang2019,Wu2020}) where RNNs are used in combination with off-policy methods, there is a substantial difference in how actions are related to the observation space. In this work, the actions taken by the agent are linear and angular accelerations, and the action space is continuous. In the observation space, which is fed in the input layer of the NN, only the heading angle is included, giving an extremely limited intuition of the state of the UAV. In the cited works, the action space has a simpler structure, designed with only one physical variable, in particular the heading angle or the acceleration, and sometimes considering a discrete action space. This implies that our algorithm not only has to find the right actions separately, but also combine them in order to have the most efficient combination of the two actions to reach the target, not forgetting to maintain a reasonable distance from the obstacles and minimizing energy consumption. Moreover, here the UAV s significantly affected by the surrounding flow-field, which is turbulent and contains several recirculation zones. In the previously cited works, the flow field from a numerical simulation is not present, thus requiring less variables to describe the behavior of the UAV in a quiescent environment. These aspects show how the present method has potential to be extended to face more complex problems, such as navigation in a 3D complex environment and be prone to be adapted to accomplish multiple tasks, for example goods delivery, with a small development effort.

\section{Conclusions and outlook}
\label{Conclusions}
Previous studies in UAV navigation tasks with RL focus on simplified or static environments, where the flow dynamics are not explicitly modeled through high-fidelity numerical simulations. The method presented in this work incorporates a flow database obtained from high-fidelity simulations of a turbulent flow-field for a navigation task in the presence of two obstacles using a PPO algorithm enhanced with LSTMs architecture. The proposed architecture puts the LSTM cell not as the input layer of the network, but after a fully connected layer which performs a first feature extraction. The first layer focuses on pattern extraction, while the LSTM focuses on modeling temporal dependencies. Moreover, this makes the architecture suitable to learn from relatively sparse data and deal with the stochasticity of the environment and observation space. \\
The algorithm was compared with a simple PPO and a TD3 architecture, showing a significant improvement in the final reward and success rate in reaching the target. PPO+LSTM architecture reached the highest reward (40.15) whereas TD3 and PPO reached a value of 18.99 and 21.15, respecgtively. Moreover, the agent learned not only to reach the target, but to reach it safely and exploiting the features of the flow field, saving energy and avoiding unnecessary linear and angular accelerations. \\
The next step will be the testing of the algorithm in a 3D environment, possibly extending the model of the UAV to a real-body problem, including forces acting on it during navigation and introducing obstacles of different heights and distances. We will also focus on and also reducing the noise produced by the drone in a real application, which would increase the level of acoustic pollution in cities.

\section*{Appendix}
\subsection*{Policy-gradient Methods}
Policy-gradient methods parameterise and optimize policies by maximizing the expected returns explicitly, unlike value-based methods that derive policies indirectly by a value function. \\
In policy-gradient methods, the expected cumulative reward has to be maximized as:
\begin{equation}
    J(\theta) = \mathbb{E}_{\pi_{\theta}}\Bigg[\sum_{t=0}^{T}\gamma^{t}r(s_t,a_t)\Bigg],
\end{equation}
where $\gamma^t$ is the discount factor raised to the power of  $t$, which discounts the reward  $r(s_t, a_t)$  received at time $t$. \\
The policy-gradient theorem yields the gradient of the objective function calculated with respect to the policy characterized by the set of weights of the NN $\theta$ (\cite{Sutton2018}):
\begin{equation}
    \nabla_{\theta}J_{\theta}=\mathbb{E}_{\pi_{\theta}}[\nabla_{\theta}\textrm{log}\pi_{\theta}(a|s)Q^{\pi_{\theta}}(s,a)],
\end{equation}
with $Q^{\pi_{\theta}}(s,a)$ being the action-value function under the policy $\pi_{\theta}$. \\
In this work we use a stochastic policy in the training steps. The output of the policy will be the covariance ${\sigma}_{\theta}$ and the mean value ${\mu}_{\theta}$ of the  Gaussian distribution.

\subsection*{Proximal-Policy Optimization}
The proximal-policy optimization (PPO) is an advanced policy-gradient method which uses a first-order trust region to limit the policy update step size and prevent too large variations of the policy, balancing at the same time exploration and exploitation (\cite{schulman2017}). This is achieved by clipping to zero the probability that the new policy $\pi_{\theta}$ deviates more than a given $\epsilon$ from the previous policy $\pi_{\theta_{\textrm{old}}}$ as follows:
\begin{equation}
\begin{split}    
     J^{\textrm{clipped}}(\theta)= \mathbb{E}_{\tau\sim\pi_{\theta}}\Bigg[\frac{1}{K}\sum_{k=0}^{K-1}\textrm{min}(\Tilde{r}_{k}(\theta)A^{\pi_{\theta}}(\boldsymbol{y}_{k},\boldsymbol{u}_{k}), \\
     \textrm{clip}(\Tilde{r}_{k}(\theta), 1-\epsilon,1+\epsilon)A^{\pi_{\theta}}(\boldsymbol{y}_{k},\boldsymbol{u}_{k}))\Bigg],
\end{split}
\end{equation}
with:
\begin{equation}
    \Tilde{r}_{k}(\theta) = \frac{\pi_{\theta}(\boldsymbol{u}_{k}|\boldsymbol{y}_{k})}{\pi_{\textrm{old}}(\boldsymbol{u}_{k}|\boldsymbol{y}_{k})},
\end{equation}
\\
being the ratio between the new and the old policy and: 
\\
\begin{equation}
    A^{\pi_{\theta}}(\boldsymbol{y}_{k},\boldsymbol{u}_{k})=Q^{\pi_{\theta}}(\boldsymbol{y}_{k},\boldsymbol{u}_{k})-V^{\pi_{\theta}}(\boldsymbol{y}_{k}),
\end{equation}
the advantage function. $\epsilon$ is the clip range, which prevents dramatic changes in the policy. \\
The advantage function represents the average improvement in the cumulative reward by selecting a specific action. It is usually computed using an estimate of the value function $V_{\phi}$ by the generalized advantage estimator (GAE) (\cite{schulman2018}):
\begin{equation}
    A_{\phi,k}=\sum_{k'=k}^{K-1}(\gamma \lambda)^{k'-k}\delta_{\phi,k'},
\end{equation}
where
\begin{equation}
    \delta_{\phi,k}=R_{k}+\gamma V_{\phi, k+1}-V_{\phi, k}
\end{equation}
is the temporal difference residual of $V_{\phi,k}$ with a discount $\gamma$ and $\lambda\in[0,1]$ is the GAE factor. \\ 
The mean-squared error (MSE) $L^{\textrm{value}}(\phi)$ between the value function estimate $V_{\phi}$ and the cumulative reward calculated from step $k$ it is the loss function used to update the critic network:
\begin{equation}
    L^{\textrm{value}}(\theta,\phi) = \mathbb{E}_{\tau\sim\pi_{\theta}}\Bigg[\frac{1}{K}\sum_{k=0}^{K-1}\Bigg(V_{\phi,k}-\sum_{k'=k}^{K}R_{k'}\Bigg)^{2}\Bigg],
\end{equation}
If the value function estimate has both the actor and critic roles, being then an additional output of the policy network, $\phi=\theta$ and the MSE is added to the clipped objective function to get the final PPO:
\begin{equation}
    J^{\textrm{tot}}(\theta)=J^{\textrm{clipped}}(\theta)-\lambda_{v}L^{\textrm{value}}(\theta),
\end{equation}
with $\lambda_{v}$ being the value-function coefficient, which measures the weight of the last term in the total objective function. \\
The steps of the algorithm are the following: first an initial policy $\pi_{\theta}$ and a value function $V_{\phi}$ are initialized. Then the interaction with the environment starts to gather a set of trajectories $\left\{s_t,a_t,r_t,s_{t+1}\right\}$ and the advantage function is estimated using GAE to enhance stability and reduce the variance of the policy throughout the training. The clipped objective function is then used to update the policy parameters and the value function parameters are adjusted to minimize the MSE. This process is then repeated until convergence of the policy is achieved, meaning that the reward does not oscillate anymore, so that the calculated standard deviation stays withing acceptable bounds and the variance reduces over time.

\subsection*{Integration of LSTM newtowrks with PPO}
The effectivenss of PPO can be enhanced by implementing LSTM cells, especially when dealing with environments in which sequential data and temporal dependencies are crucial, as it is in trajectory-optimization problems (\cite{Hochreiter1997}). LSTMs are recurrent neural networks (RNNs) that handle sequences and capture long-term dependencies, making them suitable for problems which require memory of previous states. \\
The drawbacks of traditional RNNs are addressed by LSTM cells which can keep information over long periods. LSTM cells have three main components: a forget gate, an input gate and an output gate. The forget gate decides which information has to be discarded from the cell state, and it is defined as:
\begin{equation}
    f_{t}=\sigma(W_{f}\cdot[h_{t-1}, x_{t}]+b_{f}),
\end{equation}
with $\sigma$ is the sigmoid function, $W_{f}$ are the weights, $h_{t-1}$ the previous hidden state, $x_{t}$ the current input and $b_{f}$ the bias, respectively. \\
The input gate provides the new information which has to be added to the cell state. It is defined as: 
\begin{equation}
    i_{t}=\sigma(W_{i}\cdot[h_{t-1}, x_{t}]+b_{i}),
\end{equation}
and the candidate cell state $\Tilde{C}_{t}$ can be expressed as:
\begin{equation}
    \Tilde{C}_{t} = \textrm{tanh}(W_{C}\cdot[h_{t-1}, x_{t}]+b_{C})
\end{equation}
The new cell state is then updated as:
\begin{equation}
    C_{t}=f_{t}\odot C_{t-1}+i_{t}\odot\Tilde{C}_{t},
\end{equation}
where $\odot$ indicates an element-wise multiplication. \\
The output gate controls the output based on the cell state:
\begin{equation}
    o_{t}=\sigma(W_{o}\cdot[h_{t-1}, x_{t}]+b_{o}),
\end{equation}
\begin{equation}
    h_{t}=o_{t}\odot \textrm{tanh}(C_{t}),
\end{equation}
where $o_{t}$ denotes the output state and $h_{t}$ the new hidden state respectively. \\
The combination of these components enables LSTMs to effectively capture temporal dependencies and mitigate the vanishing-gradient problem common in standard RNNs.\\
When applying LSTMs to PPO, these cells have to be integrated both in the policy and value networks. This allows the agent to learn also from past experiences, which is crucial when dealing with POMDPs. In fact, many real-world scenarios are not represented by a fully observable state of the environment. LSTMs enhance the agents' behavior by keeping information from past observations and the agents can then infer missing information and take actions based on more informed decisions. \\
Combining LSTM networks with PPO gives several advantages:
\begin{itemize}
    \item The agent can learn from temporal sequences, leading to a better-informed decision-making process;
    \item LSTMs make the whole process more suitable for complex environments with partial observability, leveraging the long-term dependencies they are able to learn;
    \item The agent can store and reuse past experiences and this a key feature for problems which are characterized by temporal dependencies.
\end{itemize}
On the other hand, they exhibit some drawbacks. In particular, the computational cost increases, they require a careful hyperparameter tuning and stabilizing the training can be more challenging due to the interaction between PPO and LSTM networks.

\section*{Acknowledgments}
Federica Tonti and Ricardo Vinuesa acknowledge funding from the European Union’s HORIZON
Research and Innovation Program, project REFMAP, under Grant Agreement number 101096698. The computations were carried out at the supercomputer Dardel at PDC, KTH, and the computer time was provided by the
National Academic Infrastructure for Supercomputing in Sweden (NAISS).
The authors also want to thank Luca Biferale and Michele Buzzicotti for their contribution by providing the data for the reproduction and validation of the Zermelo's problem, which gave foundations to the present work.

\section*{Data Availability Statement}
All the codes and data used in this work will be made available open access when the article is published here: \url{https://github.com/KTH-FlowAI}

\bibliography{biblio}

\end{document}